\documentclass{article}

\usepackage{arxiv}

\usepackage[utf8]{inputenc} 
\usepackage[T1]{fontenc}    
\usepackage{hyperref}       
\usepackage{url}            
\usepackage{booktabs}       
\usepackage{amsfonts}       
\usepackage{nicefrac}       
\usepackage{microtype}      
\usepackage{lipsum}		
\usepackage{graphicx}
\usepackage{natbib}
\usepackage{doi}
\usepackage{caption}

\title{Challenge Results Are Not Reproducible}

\author{ {Annika Reinke}\thanks{Shared first authors.} \\
	Intelligent Medical Systems (IMSY) and HI Helmholtz Imaging, 
 German Cancer Research Center (DKFZ) \\
    Heidelberg University \\
	\texttt{a.reinke@dkfz.de} \\
	\And
	{Georg Grab$^*$} \\
	Intelligent Medical Systems (IMSY),
    German Cancer Research Center (DKFZ) \\
    Heidelberg University \\
    \And
	{Lena Maier-Hein} \\
	Intelligent Medical Systems (IMSY) and HI Helmholtz Imaging, 
    German Cancer Research Center (DKFZ) \\
    Heidelberg University \\
}

\renewcommand{\shorttitle}{Challenge Results Are Not Reproducible}

\hypersetup{
pdftitle={Challenge Results Are Not Reproducible},
pdfauthor={Annika Reinke, Georg Grab, Lena Maier-Hein},
}

\begin{document}
\maketitle

\begin{abstract}
	While clinical trials are the state-of-the-art methods to assess the effect of new medication in a comparative manner, benchmarking in the field of medical image analysis is performed by so-called challenges. Recently, comprehensive analysis of multiple biomedical image analysis
challenges revealed large discrepancies between the impact of challenges and quality control of the design and reporting standard. This work aims to follow up on these results and attempts to address the specific question of the reproducibility of the participants methods. In an effort to determine whether alternative interpretations of the method description may change the challenge ranking, we reproduced the algorithms submitted to the 2019 Robust Medical Image Segmentation Challenge (ROBUST-MIS). The leaderboard differed substantially between the original challenge and reimplementation, indicating that challenge rankings may not be sufficiently reproducible.
\end{abstract}


\section{Introduction}
\label{3341-introduction}
Robust segmentation of biomedical images is an important precursor to many new, innovative computer-assisted applications. Deep learning-based segmentation methods have proven to work successfully on a wide range of medical imaging data, including computed tomography (CT), magnetic resonance imaging (MRI), and endoscopy \cite{3341-01}. For benchmarking which type of model works best on a given medical domain, challenges have become an important tool, and are now commonplace in conferences such as the conference on Medical Image Computing and Computer Assisted Interventions (MICCAI) or the  IEEE International Symposium on Biomedical Imaging (ISBI). However, recent comprehensive analysis of challenges in the biomedical domain revealed that the current state of quality control severely limits interpretation of rankings and reproducibility, with only a fraction of the relevant information typically provided \cite{3341-02}.

In order to concretely analyze the reproducibility of the participating methods in challenges, we aimed to reimplement the algorithms of all participating teams in a challenge only based on their submitted method descriptions. As an example, we performed our experiments for the 2019 Robust Medical Image Segmentation Challenge (ROBUST-MIS). Given the obligation to submit a detailed description of their methods together with their actual results, this challenge had a disproportionately high amount of algorithmic information available, which should in theory faciliate the reproducibility of results. However, in this work, we show that even with this high amount of information available, we were not able to reproduce the challenge results.

\section{Materials and Methods}
\label{3341-methods}
The ROBUST-MIS challenge \cite{3341-03} focused on the robustness and generalization capabilities of algorithms. A collection of surgical data with 10 040 annotated images from 30 surgical procedures across three different types of surgery served as the basis for the challenge. The challenge was validated across competing methods in three stages with a growing domain gap between the training and test data, i.e. higher stages contained more difficult images requiring a higher degree of generalization to be segmented successfully. A detailed overview of the challenge can be found in \cite{3341-03}. In the following experiments, we focused on the multi-instance instrument segmentation task of the challenge.

In the challenge, alongside their algorithm submission, participating teams were required to submit a document summarizing their method in detail to the point of being reproducible, such as the used network architecture, data augmentations and all hyperparameters. These method descriptions, along with the summaries included in the challenge paper \cite{3341-03} were used as a basis for reproducing the challenge results. In general, we aimed to stay as close to the descriptions as possible, meaning the same programming languages and libraries were used, if this information was made available. 

In case of ambiguous or missing information in method descriptions, we first attempted to infer the correct meaning using literature directly cited by the method description. Only if this was not possible, secondary literature was considered. As a last resort, we filled the missing information by surveying publicly available similar implementations and taking the most popular approach that worked reasonably well on the problem domain. For example, if a team would not document the type of optimizer, and relevant citations did not explicitly mention this either, the default choice of the most popular or official implementation was used. If two interpretations were equally likely, the method was trained using both interpretations, and the one resulting in better validation performance was chosen. 

In the original challenge, participants were ranked according to two different criteria, robustness and generalization capabilities, resulting in two rankings based on the multi-instance Dice Similarity Coefficient (MI\_DSC) \cite{3341-03, 3341-04}. The robustness ranking was determined by calculating a metric-based ranking using the 5\% quantile of MI\_DSC values obtained from the testing set. The accuracy ranking was calculated as a test-based ranking using a Wilcoxon signed rank test at a 5\% significance level \cite{3341-05}. For our calculation, we considered stage three of the test set. Additionally, we compared the rankings with Kendall's $\tau$ correlation coefficient \cite{3341-06}, which yields a value of 1 for two perfectly agreeing rankings and -1 if rankings are reversed. Ranking variability was investigated via bootstrapping \cite{3341-05}. We used the \emph{challengeR} package \cite{3341-05} for calculating rankings and ranking uncertainty.

\section{Results}
\label{3341-results}
\begin{figure}[h]
    \centering
    \includegraphics[width=0.85\textwidth]{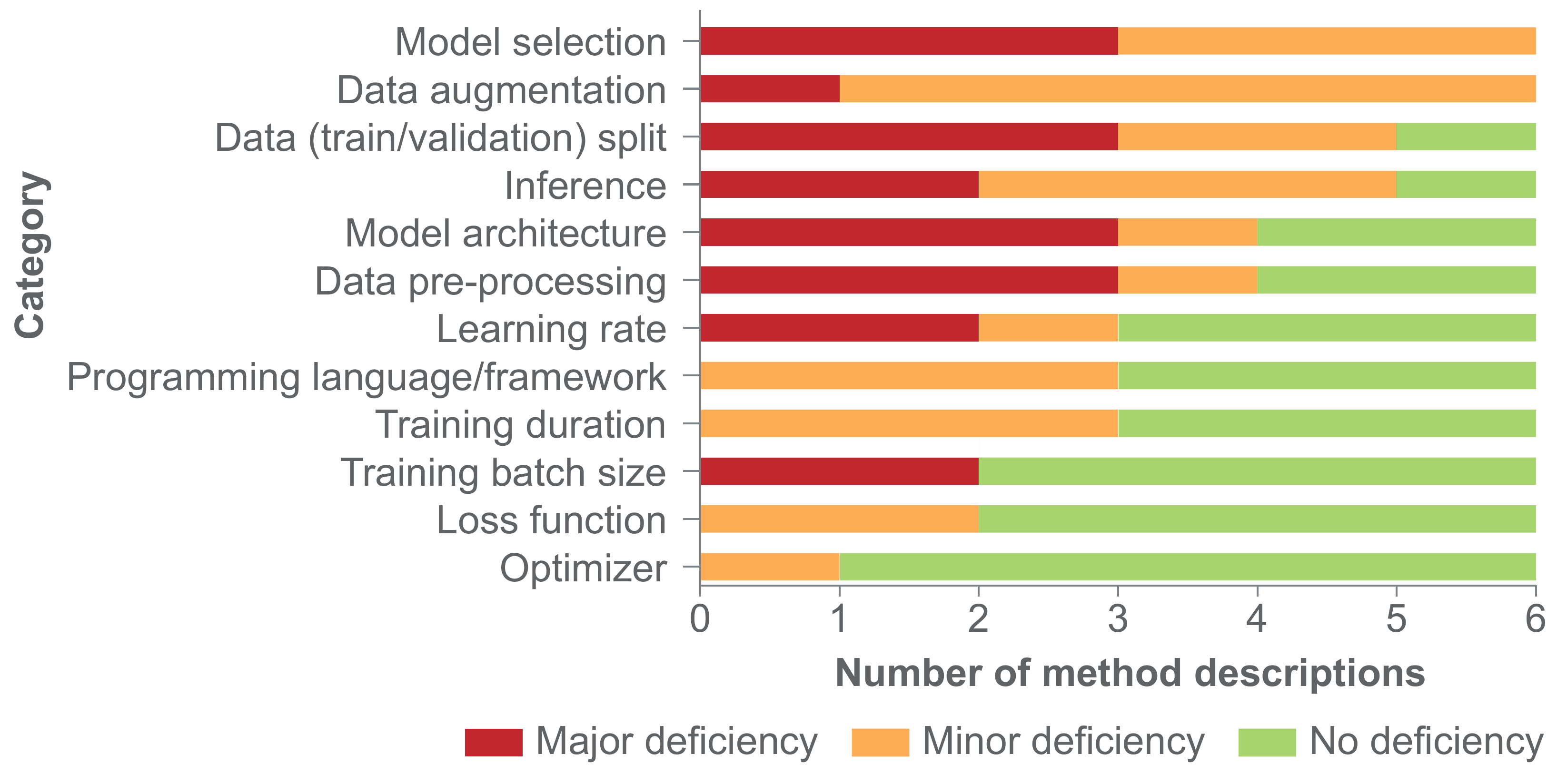}
    \caption{Qualitative analysis of deficiency in method descriptions across several different aspects of implementation.}
    \label{3341-fig1}
\end{figure}

During our reimplementation, lots of ambiguities were found in the method descriptions. Fig. \ref{3341-fig1} presents a qualitative summary of the assumptions made across all descriptions. Here, the term minor deficiency was defined as an assumption that had to be taken due to missing or clearly incorrect information, but was thought to either have a minor impact on model performance or there was high confidence that the right assumption has been made from context. Major deficiencies were defined as missing design decisions either thought to have a major impact on final model performance, there was low confidence that the correct assumption had been made from context or context was unavailable. In such a case, it was highly unlikely that our choice was identical to that of the original implementation. From the figure, it can be seen that both the model selection and data augmentation showed the highest amount of major and minor deficiencies during the reimplementation, followed by the data splits and the description of inference.

\begin{table}[t]
    \caption{Accuracy Ranking based on the multi-instance Dice Similarity Coefficient (MI\_DSC). Accuracy is determined by the proportion of significant tests divided by the number of algorithms (Prop. Sign). The last column of (b) $\Delta$ shows the relative rank difference between the original challenge and the reimplementation of the algorithm.}
    \begin{minipage}[h]{0.45\textwidth}
        \centering
        \setlength{\tabcolsep}{0.5em}
        \begin{tabular}{ccc}
        & &\\
        \hline
       Team & Prop. Sign. & Rank\\
        \hline
        A1 & 1.00 & 1\\
        A2 & 0.83 & 2\\
        A3 & 0.67 & 3\\
        A4 & 0.33 & 4\\
        A5 & 0.33 & 4\\
        A6 & 0.17 & 6\\
        A7 & 0.00 & 7\\ \hline
       \end{tabular} \\
       (a) Original
       \label{3341-tab-accuracy-ranking_original}
    \end{minipage}
    \hfill
    \begin{minipage}[h]{0.45\textwidth}
        \centering
        \setlength{\tabcolsep}{0.5em}
        \begin{tabular}{cccc}
        & &&\\
        \hline
       Team & Prop. Sign. & Rank & $\Delta$ \\
        \hline
        A2 & 1.00 & 1 & ↑ 1 \\
        A1 & 0.83 & 2 & ↓ 1 \\
        A4 & 0.67 & 3 &  ↑ 1\\
        A5 & 0.50 & 4 & $\rightarrow$ 0\\
        A3 & 0.33 & 5 & ↓ 2 \\
        A7 & 0.17 & 6 & ↑ 1\\
        A6 & 0.00 & 7 & ↓ 1\\ \hline
       \end{tabular} \\
       (b) Reimplementation
       \label{3341-tab-accuracy-ranking_reimplementation}
    \end{minipage}
    \label{3341-tab-accuracy-ranking}
\end{table}
When calculating the metric values of the reimplemented methods, the distribution of values substantially differed between the original challenge and the reimplementation, except for team A2. This was also visible in the rankings. Tab. \ref{3341-tab-accuracy-ranking} shows the accuracy ranking for the original challenge and the reimplementation. The original winner changed for the reimplementation and teams moved mostly up or down by one single rank with an average change of one rank. Kendall's $\tau$ was 0.59 between both rankings, indicating a high variability. The ranking variability was analyzed by applying bootstrapping. The average (median, Interquartile Range (IQR) Kendall's $\tau$ over 1,000 bootstrap rankings was 1.00 (median: 1.00; IQR: (1.00, 1.00)) for the original challenge, which was thus very robust against small perturbations. The average (median, IQR) Kendall's $\tau$ for the reimplementation was slightly less with a mean (median, IQR) Kendall's $\tau$ of 0.98 (median: 0.98; IQR: (0.98, 1.00)).

\begin{table}[t] 
		\centering
  \caption{Robustness rankings based on the 5\% quantile of the multi-instance Dice Similarity Coefficient (MI\_DSC), computed for all stage 3 test cases. The last column of (b) $\Delta$ shows the relative rank difference between original challenge and reimplementation for the algorithm.}
    	\label{3341-tab-robustness-ranking}
        \begin{minipage}[h]{0.45\textwidth}
        \centering
            \setlength{\tabcolsep}{0.5em}
            \begin{tabular}{ccc}
            & &\\
            \hline
           Team & MI\_DSC & Rank \\
            \hline
            A5 & 0.31 & 1\\
            A2 & 0.26 & 2\\
            A4 & 0.22 & 3\\
            A7 & 0.19 & 4\\
            A1 & 0.17 & 5\\
            A3 & 0.00 & 6\\
            A6 & 0.00 & 6\\
            \hline
           \end{tabular} \\
           (a) Original
           \label{3341-tab-robustness-ranking-original}
        \end{minipage}
        \hfill
        \begin{minipage}[h]{0.45\textwidth}
        \centering
            \setlength{\tabcolsep}{0.5em}
            \begin{tabular}{cccc}
            & &&\\
            \hline
           Team & MI\_DSC & Rank & $\Delta$ \\
            \hline
            A2 & 0.28 & 1 & ↑ 1\\
            A1 & 0.11 & 2 &  ↓ 1\\
            A5 & 0.04 & 3 & ↓ 2\\
            A3 & 0.00 & 4 & ↑ 2\\
            A7 & 0.00 & 4 & $\rightarrow$ 0\\
            A4 & 0.00 & 4 & ↓ 1\\
            A6 & 0.00 & 4 & ↑ 2\\ 
            \hline
           \end{tabular} \\
           (b) Reimplementation
           \label{3341-tab-robustness-ranking-reimplementation}
        \end{minipage}

\end{table}

Similarly, Tab. \ref{3341-tab-robustness-ranking} shows the original and reimplemented versions for the robustness ranking. Again the winners according to this ranking changed and the average change in ranks was higher for this ranking scheme (1.3). Comparing both rankings yielded a Kendall's $\tau$ of 0.40. Notably, four algorithms failed to achieve a 5\% quantile of the MI\_DSC above $0$ in the reimplementation, which only happened for two algorithms in the original challenge. We further found a higher ranking uncertainty for the original challenge with a mean (median, IQR) Kendall's $\tau$ of 0.85 (median: 0.98; IQR: (0.98, 1.00)). On the other hand, this ranking scheme was more stable for the reimplementation (mean: 0.97; median: 1.00; IQR: (1.00, 1.00)).

\section{Discussion}
\label{3341-discussion}
In this work, we attempted to reproduce the rankings of the ROBUST-MIS challenge by means of reimplementing algorithms of participating teams given the method description they were required to submit. This attempt failed: both ranking schemes yielded results that substantially differed for the reimplementation, including changing the winners. 

While training deep learning models comes with a substantial amount of non-determinism, which additionally contributes to the problem of reproducibility, we think the primary reason for failing to reproduce the results is the insufficient documentation provided by the participants. As shown in Fig.~\ref{3341-fig1}, the number of assumptions needed to be taken for reproducing the methods were numerous and spanned all relevant steps of model development, including data preprocessing, model architecture, and inference. For one team, we were not even able to identify the basic network architecture.

We found that complex design decisions tended to be described less accurately than design decisions that are typically simpler to document. For example, the standard choices for optimizers are limited and typically prominently visible in the source code. This may be a reason why almost all participants succeeded in unambiguously stating the utilized optimizer and associated hyperparameters. 

On the other hand, model selection and data augmentation are complex processes, which were documented poorly by challenge teams. While the best-performing model is usually selected by calculating the loss on a separate validation data set, this does not necessarily have to be the case. In the ROBUST-MIS challenge, in particular, it was beneficial to select a sensitive over a specific model, since a false negative fraction of only 5\% would be enough to completely fail the robustness ranking, i.e. yielding a 5\% quantile of $0$. Many teams either overlooked this aspect of the challenge completely in their documentation or provided incomplete information. Similarly, while the types of data augmentations were typically well reported, the respective hyperparameters were usually not documented. In addition, data augmentations can be applied individually or be combined with other data augmentation techniques. In such a case, the order and probabilities need to be specified. Finally, data augmentations complicate the exact meaning of the term 'epoch': is the original dataset extended only once with a certain percentage of augmented images, or are augmentations continually applied on the fly during training? All these choices need to be documented in detail in order to allow for faithful reimplementation.

Most design choices going into an algorithm relevant for challenge participation directly map to the source code, and thus reproducibility would be greatly improved by making the source code publicly available. However, since this is practically challenging, e.g. for teams from industry, certain aspects of the method description should be handled with great care:

\emph{Reasoning for complexity:} Some teams made complicated design decisions. For example, one team used a complex multi-stage approach for inference but did not elaborate on the reasoning for choosing this procedure. While a detailed explanation would have increased the understanding in general, it could also have been used to verify that an implementation was correct while reproducing the results.

\emph{Hyperparameters:} Although simple to document, many teams failed to properly list their chosen hyperparameters, especially for data augmentation and final threshold values for the purpose of inference.

\emph{Model Selection:} While most design decisions directly map to the source code, model selection is often a notable exception to this, and may involve manual analysis and comparison of several models using different performance metrics. This may be a reason why this work identified many deficiencies related to this aspect. Especially in segmentation tasks, the considerations may go beyond minimizing the validation loss, since the final ranking methods are often not suitable for being utilized as loss functions. In any case, model selection should ideally be quantifiable and documented. 

It should be noted that drawing conclusions from this work is limited since only a single challenge has been analyzed. However, for this challenge, an exceptionally high amount of information regarding the algorithms was available, strengthening our hypothesis that reproduction of challenge results is limited even if a detailed method description is required from the organizers. Furthermore, training deep learning models is inherently associated with a certain degree of non-determinism, where two identical training runs can potentially lead to severely different results \cite{3341-07}. Only one challenge participant addressed this limitation by employing ensembling and averaging their results during inference. Thus, ironically, this work itself may be deemed non-reproducible. 

With this work, we showed that even well-documented methods are not easily reproducible. However, we think that the most effective way of reducing the issue of non-reproducibility would be publicly available source code of all participating teams of a challenge, although maybe practically challenging. Especially for the winning teams, such an action would be desirable since the winning method is typically seen as the new state-of-the-art method for a specific problem. We hope that this work will trigger further actions by stakeholders involved in policy-making for challenges.

\section*{Acknowledgements}
Part of this work was funded by Helmholtz Imaging, a platform of the Helmholtz Incubator on Information and Data Science. We would like to thank Marcel Knopp and Minu D. Tizabi for proofreading the document.

\bibliographystyle{unsrtnat}
\bibliography{references}  






\end{document}